\documentclass{article}

     \PassOptionsToPackage{numbers, compress}{natbib}

  \usepackage[dblblindworkshop, final]{neurips_2026}

\usepackage[utf8]{inputenc} %
\usepackage[T1]{fontenc}    %
\usepackage{hyperref}       %
\usepackage{url}            %
\usepackage{booktabs}       %
\usepackage{amsfonts}       %
\usepackage{nicefrac}       %
\usepackage{microtype}      %
\usepackage{fontawesome5}   %
\usepackage{xcolor}         %
\definecolor{citeblue}{HTML}{1A5FB4}   %
\definecolor{linkred}{HTML}{A51C30}    %
\definecolor{urlteal}{HTML}{0F766E}    %
\usepackage{graphicx}
\usepackage{subcaption}
\usepackage{wrapfig}
\hypersetup{
  colorlinks=true,        %
  citecolor=blue,     %
  linkcolor=red,      %
  urlcolor=urlteal,       %
  linktoc=all,
  breaklinks=true,
  pdfborder={0 0 0}       %
}
\usepackage{amsmath}
\usepackage[noabbrev]{cleveref}
\usepackage{float}

\newif\ifshowcomments
\showcommentsfalse

\ifshowcomments
  \newcommand{\simon}[1]{\textcolor{orange}{[Simon: #1]}}
  \newcommand{\klemens}[1]{\textcolor{violet}{[Klemens: #1]}}
  \newcommand{\leo}[1]{\textcolor{green!60!black}{[Leo: #1]}}
  \newcommand{\frank}[1]{\textcolor{cyan!60!black}{[Frank: #1]}}
  \newcommand{\eliott}[1]{\textcolor{gray!60!black}{[Eliott: #1]}}
\else
  \newcommand{\simon}[1]{\ignorespaces}
  \newcommand{\klemens}[1]{\ignorespaces}
  \newcommand{\leo}[1]{\ignorespaces}
  \newcommand{\frank}[1]{\ignorespaces}
  \newcommand{\eliott}[1]{\ignorespaces}
\fi

\title{Out-of-the-Box Tabular Foundation Models for Bayesian Optimization in Materials Discovery}

\workshoptitle{AI4Mat}
\hypersetup{
  pdftitle={Out-of-the-Box Tabular Foundation Models for Bayesian Optimization in Materials Discovery},
  pdfauthor={Klemens Fl\"oge, Simon Bing, Leo Grinsztajn, Eliott Kalfon, Lennart Purucker, Noah Hollmann, Frank Hutter}
}

\author{%
  Klemens Fl\"oge$^{1}$ \quad Simon Bing$^{1}$ \quad Leo Grinsztajn$^{1}$ \quad Eliott Kalfon$^{1}$ \\[3pt]
  \bfseries Lennart Purucker$^{1,2}$ \quad Noah Hollmann$^{1}$ \quad Frank Hutter$^{1,2,3}$ \\[5pt]
  \normalfont\small $^{1}$Prior Labs \quad  $^{2}$University of Freiburg\\
  \normalfont\small $^{3}$ELLIS Institute T\"ubingen
}

\begin{document}

\maketitle

\begin{abstract}
Bayesian optimization is widely used in active learning loops for materials science. Gaussian processes serve as the de-facto surrogate model family, but require task-dependent kernel choices and can lack the modeling capacity for complex relationships. Tabular foundation models (TFMs) have recently emerged as the state-of-the-art for predictive tasks on structured data, without requiring bespoke kernels or hyperparameter tuning. We evaluate two TFMs, TabPFN and TabICL, as out-of-the-box surrogate models on two standard materials science benchmarks, comparing against a large field of baselines encompassing Gaussian processes, random forests, and CatBoost. Our findings indicate that TFMs are highly competitive surrogate models across the two benchmarks, without task-specific training for active learning or GP kernel selection. Our results motivate developing TFMs further as surrogates for active learning in materials discovery.
\par\medskip
\centerline{\hypersetup{urlcolor=blue}\faGithub\ \url{https://github.com/PriorLabs/matsci-bo-bench}}
\end{abstract}

\section{Introduction}
Materials discovery is characterized by costly experimental pipelines, where rapid iteration cycles are stifled by the cost of evaluating a candidate set of input parameters. As such, Bayesian optimization (BO) is a natural paradigm to explore the space of experimental configurations, given past results \citep{gongora2020bayesian,mekkiberrada2021twostep,deneault2021toward}. As running a full active learning campaign for each new candidate method is both impractical and ineffective, methods are commonly evaluated and compared retrospectively in completed learning campaigns \citep{rohr2020benchmarking,liang2021benchmarking}. The default surrogate model for the BO loop in these settings is a Gaussian process (GP) \citep{liang2021benchmarking,hase2021olympus,shields2021bayesian}.

However, \citet{muller2023pfns4bo} have shown in-context learning to be a general and flexible method for surrogate modeling, with strong performance in hyperparameter optimization. While this approach relies on task-dependent priors provided by the user, Tabular Foundation Models (TFMs), trained on a single task-independent prior, are promising candidates as general surrogates for BO. Indeed, these models have become the state of the art for predictive tasks on structured data \citep{erickson2025tabarena, purucker2026iidgeneraltabularfoundation}, by leveraging the principle of pretraining on large-scale synthetic datasets in place of per-dataset fitting \citep{hollmann2025tabpfn,qu2026tabiclv2,grinsztajn2025tabpfn25, grinsztajn2026tabpfn3technicalreport, zhang2025mitra, zhang2025limix}. TFMs thus combine native predictive distributions \citep{landsgesell2026scoringbenchbenchmarkevaluatingtabular} with the modeling capacity of neural network-based surrogates \citep{springenberg2016bayesian}, without requiring campaign-specific kernel selection.

In practice, GIT-BO~\citep{yu2026gitbo} utilizes the TFM TabPFN in a high-dimensional continuous engineering design setting, and \citet{rogers2026zeroshot} find an off-the-shelf TabPFN competitive with BO-pretrained transformers on hyperparameter tuning. None of these settings is materials discovery, and rankings need not carry across domains \citep{yu2026bocode}.
\citet{hu2026foundation} provide first evidence that TFMs can outperform GPs in small-budget chemistry campaigns, under their own protocol, against one GP, and with a Gaussian approximation of TabPFN's uncertainty. We instead use published benchmark protocols, seven GP kernels and the TFMs' native predictive distributions (\Cref{sec:hu}). We share their conclusion, not their finding that TabPFN needs $10$-$20\%$ of the pool to overtake a GP (\Cref{fig:ni-by-budget}). %

\begin{figure}[t]
  \centering
  \includegraphics[width=\linewidth]{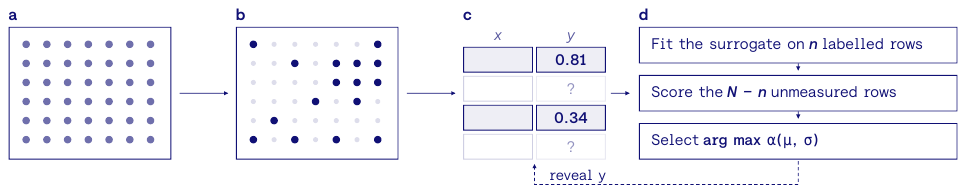}
  \caption{\textbf{Retrospective, pool-based BO.} \textbf{a}, candidate conditions in the experimental design space. \textbf{b}, the conditions the original campaign measured form the pool. \textbf{c}, the optimizer sees the features $x$ of every candidate; outcomes $y$ stay hidden until queried. \textbf{d}, each round, the surrogate is fit on the queried rows and scores the rest; the acquisition rule picks the next candidate, and its recorded outcome is revealed.}
  \label{fig:intro_diagram}
\end{figure}

\paragraph{Scope and contribution.} We ask whether off-the-shelf TFMs, with no training for BO or chemistry and no tuning, work as drop-in surrogates under the published protocols of established materials benchmarks; we propose no new method. We evaluate TabPFN and TabICL on 15 retrospective, pool-based campaigns from two benchmarks \citep{liang2021benchmarking,hase2021olympus}, against 56 arms: seven GP kernels, random forests, CatBoost and both TFMs, each under five acquisition rules, plus random search. A TabPFN arm ranks first on both benchmarks (\Cref{fig:ni_auc_bars_main}). By avoiding campaign-specific GP kernel selection, tabular foundation models offer a low-friction surrogate for small-data, closed-loop materials optimization.

\section{Setup}
\label{sec:setup}

We consider retrospective, pool-based active learning tasks and study how using TFMs as surrogate models in a Bayesian optimization (BO) loop compares to other methods. A \emph{pool} is the finite set of candidate conditions a published study actually measured, together with its recorded outcomes. An active learning algorithm explores that pool sequentially, optimizing for a given target: at each round a surrogate is fitted to what has been queried so far while an acquisition function selects the next candidate. What makes the task \emph{retrospective} is that every outcome is already on record, so a campaign can be replayed over many seeds at no experimental cost. \Cref{fig:intro_diagram} illustrates this setup.

\paragraph{Benchmark datasets.}
We use two benchmarks: the \texttt{liang} dataset containing five materials discovery tasks ($3$-$5$ variables; $94$-$600$ candidates) \citep{liang2021benchmarking} and the (raw) \texttt{olympus} datasets with ten chemistry and colorimetry tasks ($3$-$6$ variables; $66$-$1386$ candidates) \citep{hase2021olympus}. 
See \Cref{app:datasets} for all details of the benchmark datasets. 
Following \citet{liang2021benchmarking}, for each run, we randomly sample two initial candidates from the pool, acquire one more point per iteration and repeat until the pool is exhausted. Different random seeds determine the initial starting candidate pair, as well as any algorithmic randomness.
Method rankings are sensitive to the initial design, budget and scoring rule \citep{rohr2020benchmarking,borg2023quantifying}, so we adopt each benchmark's own seeds and scoring rule: 50 seeds and Top-5\% on \texttt{liang}, 100 seeds and best achieved value on \texttt{olympus}.
We deviate in one respect. \citet{hase2021olympus} fit emulators to their raw data and have planners query those emulators, whereas we replay the recorded pools directly. Selecting from a fixed candidate set is the setting we study, and an emulator smooths the design space, which may favor surrogates whose assumptions match that smoothness \citep{liang2021benchmarking}.

\paragraph{Baselines.} We pair eleven surrogates with five acquisition rules and report the full combinatorial set. As surrogate models we consider Gaussian processes with seven kernels \citep{liang2021benchmarking,snoek2014input}, since a GP's kernel choice is itself non-trivial and task-dependent \citep{duvenaud2013structure}, random forests \citep{breiman2001random}, CatBoost \citep{prokhorenkova2018catboost} and tabular foundation models (TabICL \citep{qu2026tabiclv2} and TabPFN-3 \citep{grinsztajn2026tabpfn3technicalreport}).
Every surrogate is run under the acquisition rules greedy \citep{death2021greed}, expected improvement (EI) \citep{mockus1978application,jones1998efficient}, probability of improvement (PI) \citep{kushner1964new} and $\mathrm{UCB}1$, $\mathrm{UCB}2$ \citep{cox1992statistical,srinivas2010gaussian}. 
Additionally, we report the baseline of randomly choosing the next candidate point. The full enumeration of all surrogate-acquisition pairs is given in \Cref{app:baselines}.

\paragraph{Features and predictive distributions.}
All surrogates see the same features: the benchmark's raw input variables ($3$--$6$ numeric variables), with no descriptors or feature engineering. Targets are negated for minimization tasks. GPs normalize inputs and standardize targets; the other surrogates take raw values. Every surrogate is refitted on the queried rows at each round; for TFMs, these rows form the context of one forward pass, with no weight updates. Acquisition uses each TFM's native predictive distribution, with no Gaussian approximation: EI, PI and UCB are computed exactly on TabPFN's bar distribution, and from TabICL's predicted quantile function. UCB$\beta$ is the predictive quantile at probability $\Phi(\beta)$, $\beta\in\{1,2\}$, which equals $\mu+\beta\sigma$ for a Gaussian. GPs, random forests and CatBoost use the Gaussian forms on their predictive mean and standard deviation, and greedy ranks by the mean (\Cref{tab:acquisitions}, \Cref{app:impl}).

\paragraph{Metrics.}
Top-5\% is the fraction of the pool's top 5\% candidates found. Given the different sizes of pools, budgets are stated as a fraction of the pool. To aggregate results over settings within a benchmark, we employ the Normalized Improvement (NI) metric (cf. \Cref{app:ni}) \citep{erickson2020autogluon}, which rescales every arm within a campaign and budget onto the range $[0, 1]$: the worst arm maps to $0$ and the best to $1$. This metric is then averaged over tasks to obtain the final aggregate score.

\section{Results}
\label{sec:results}

\begin{figure}[t]
  \centering
  \includegraphics[width=\linewidth]{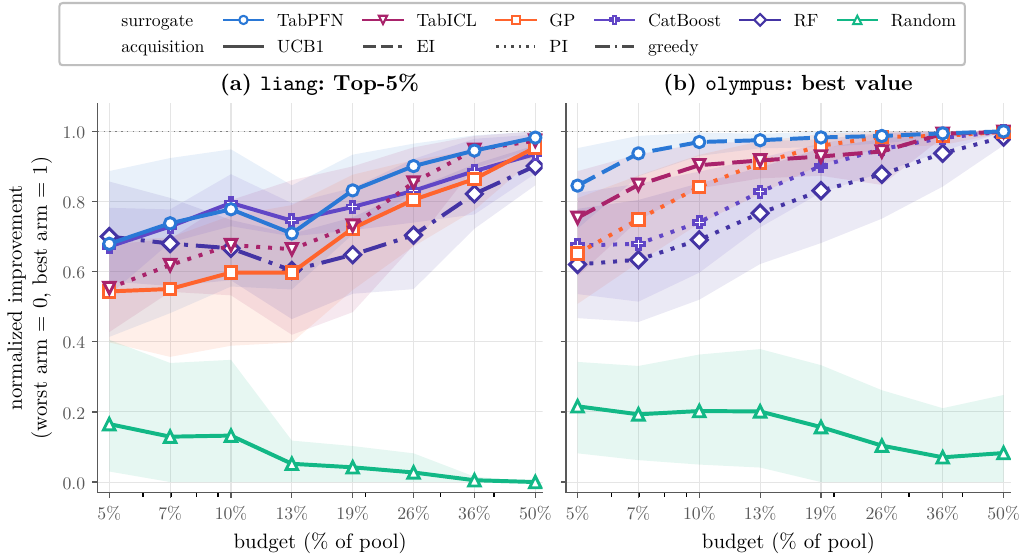}
  \caption{\textbf{Normalized improvement (NI) against budget, best arm per surrogate family.} (a)~\texttt{liang}, Top-5\%. (b)~\texttt{olympus}, best value. Bands are the $2.5$-$97.5$ percentile bootstrap over campaigns; seed-level differences within each campaign are analyzed in \Cref{app:posthoc}. TFMs are competitive across benchmarks and budgets.
  }
  \label{fig:ni-by-budget}
\end{figure}

\Cref{fig:ni-by-budget} shows the best observed arm per surrogate family, selected by area under the NI curve within each benchmark (\Cref{fig:ni_auc_bars_main}) \citep{guyon2011active,turner2021bayesian}. Selection includes the acquisition rule for every family and the kernel for GPs. These curves thus describe configurations selected on the evaluated results. All 56 arms appear in \Cref{fig:all_arms_liang_bestvalue,fig:all_arms_olympus_top5}.

Across both benchmarks, the TFMs are competitive with the other baselines over most of the budget range. Seed-to-seed spread within a campaign is large (\Cref{fig:liang_raw,fig:olympus_raw}). Following \citet{shields2021bayesian}, we test each campaign with a Welch t-test on performance AUC over seeds (\Cref{app:posthoc}). The TabPFN arm of \Cref{fig:ni-by-budget} is better than each other family's arm on most campaigns and worse on at most one (\Cref{tab:seed_tests}). Both arms are selected post hoc, so selection bias remains.

\paragraph{Sensitivity to the acquisition function.}
The best acquisition rule varies across benchmarks and individual campaigns; EI and PI lead on \texttt{olympus}, while PI and UCB$1$ are stronger on \texttt{liang}. \Cref{fig:posthoc_best_rule} shows the best observed acquisition rule for each surrogate and campaign; these campaign-specific winners differ from the benchmark-wide selection used in \Cref{fig:ni-by-budget} (\Cref{app:posthoc}). Nevertheless, the NI-AUC ranking in \Cref{fig:ni_auc_bars_main} shows that strong TabPFN performance is not confined to a single acquisition rule, with several TabPFN arms occupying top positions on both benchmarks. In particular, TabPFN-UCB$1$ and TabPFN-PI take the top two places on \texttt{liang}, and TabPFN-EI and TabPFN-UCB$2$ take the top two on \texttt{olympus}.
Thus, performance depends not only on the surrogate, but also on how it is used for acquisition.

\begin{figure}
  \centering
  \includegraphics[width=0.8\linewidth]{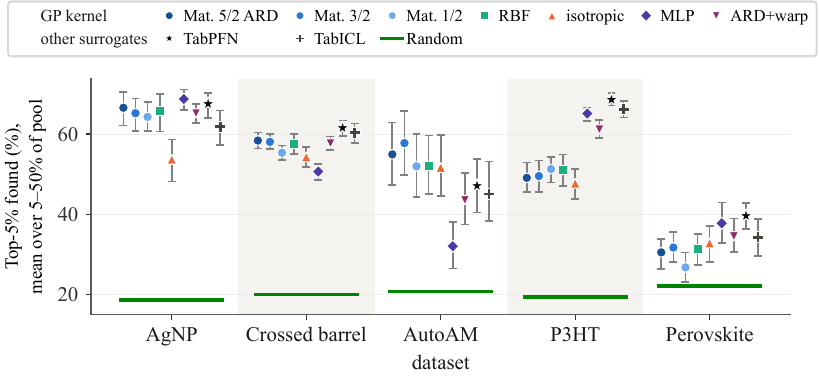}
  \caption{\textbf{GP kernel choice per dataset on \texttt{liang}.} All arms under \textsc{ucb1}, the acquisition rule of the best-ranked GP arm on \texttt{liang} (\Cref{fig:ni_auc_bars_main}). Each mark summarizes one arm's Top-5\% found over eight log-spaced budgets from 5-50\% of the pool, on the raw percentage scale.
  Grey bars are a $2.5$-$97.5$ percentile bootstrap over seeds. We report the analogous results on the \texttt{olympus} benchmark in \Cref{fig:kernels_olympus_pi}.}
  \label{fig:kernels_ucb2}
\end{figure}

\paragraph{Kernel choices for GPs.}
The per-campaign results for \texttt{liang} in \Cref{fig:kernels_ucb2} show considerable heterogeneity across GP kernels, the design of which is an active concern in chemistry applications \citep{griffiths2023gauche} and a known source of variation in this benchmark family \citep{rohr2020benchmarking,liang2021benchmarking}.
The MLP arm exemplifies this best: it performs best on some campaigns (AgNP and Perovskite), while doing worst on another (AutoAM).
On AutoAM and Perovskite, the spread across kernels is comparable to the gap to random search.
The best kernel differs across campaigns, making kernel choice a model selection problem \citep{duvenaud2013structure,malkomes2016bayesian}. Kernel misspecification can impair optimization \citep{berkenkamp2019noregret}. TFMs avoid this choice, while acquisition selection remains consequential.

\paragraph{Metric dependence.} Scoring the same runs on the alternative metric reorders the field. On \texttt{olympus}, the TFMs stand out more under the benchmark's own metric (best value) than under Top-5\%. \Cref{app:metric_flip} shows both benchmarks under the alternative metric.

\section{Discussion}
\label{sec:discussion}

We find that tabular foundation models are highly competitive surrogates for retrospective pool-based active learning in materials discovery, despite being used entirely out-of-the-box and not being trained for sequential optimization or chemistry. Across two established benchmarks with small, low-dimensional pools \citep{liang2021benchmarking,hase2021olympus}, they compete with or outperform carefully specified GP surrogates while avoiding GP kernel selection. Evaluating further benchmarks \citep{rohr2020benchmarking,shields2021bayesian} is an important next step.

\paragraph{Limitations.}
Our benchmarks are retrospective, with small pools and few variables. Replay measures how well a method reorders an existing pool, not which data it would collect, so our results do not establish superiority in live closed-loop campaigns. The pools are also shaped by earlier policies: four of five \texttt{liang} campaigns were collected along the path of a BO-guided campaign \citep{liang2021benchmarking}, potentially favoring BO-like policies. Autonomous laboratories \citep{adesiji2026benchmarking,ginsburg2023color,snapp2023driving,gamella2025causal} may enable ``live'' benchmarks.

As discussed in \Cref{sec:results}, the ranking and ultimately the interpretation of what constitutes a good model depend crucially on the chosen evaluation protocol and metric. We follow the evaluation schemes of the considered benchmarks in an attempt to minimize the effect of custom metrics, but even among the two benchmarks in this work, underlying metrics differ. The development and widespread adoption of unified benchmarks have been central to the success of TFM development \citep{erickson2025tabarena,purucker2026iidgeneraltabularfoundation}, and we believe similar efforts for materials science could catalyze the development of active learning methods as well.

\paragraph{Outlook.}

TFMs can take any auxiliary information expressible as rows, such as prior measurements, sibling campaigns or computed descriptors, directly into their context, whereas GPs typically need purpose-built multi-task or transfer kernels \citep{taylor2023accelerated}.
Existing TFMs \citep{grinsztajn2026tabpfn3technicalreport, qu2026tabiclv2} scale quadratically with context size, against the cubic cost of GPs.

Our experiments use TFMs without task-specific pretraining. Encoding domain knowledge in their synthetic training priors may improve performance further. Another direction is to adapt TFM training to sequential optimization and acquisition. Our results motivate investigating both directions for materials discovery.

\begin{ack}
We thank the entire Prior Labs team for their support, feedback, and discussions throughout this work.

We acknowledge the EuroHPC Joint Undertaking for awarding this project access to the EuroHPC supercomputer LUMI, hosted by CSC (Finland) and the LUMI consortium through a EuroHPC AI Factory Large Scale Access call (project EHPC-AIF-2025LS17-014).

We acknowledge EuroHPC JU for awarding the project ID EHPC-AIF-2025LS14-031 access to MareNostrum 5 ACC hosted by the Barcelona Supercomputing Center (BSC), Spain.

\begin{center}
\hfill
\begin{minipage}{0.3\textwidth}
\centering
\includegraphics[width=\textwidth]{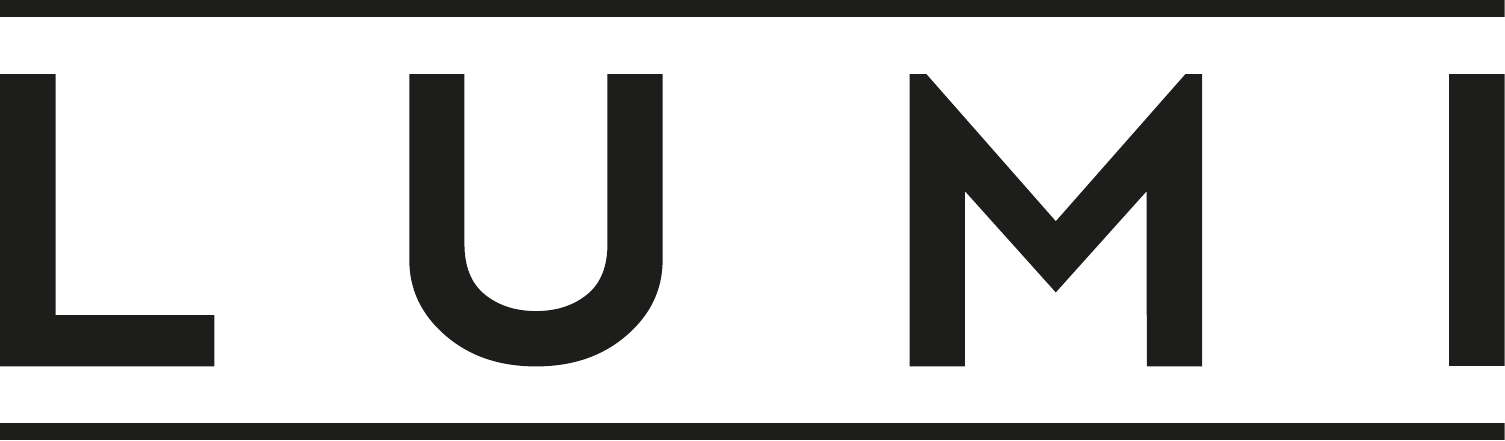}
\end{minipage}
\hfill
\begin{minipage}{0.45\textwidth}
\centering
\includegraphics[height=1.3cm]{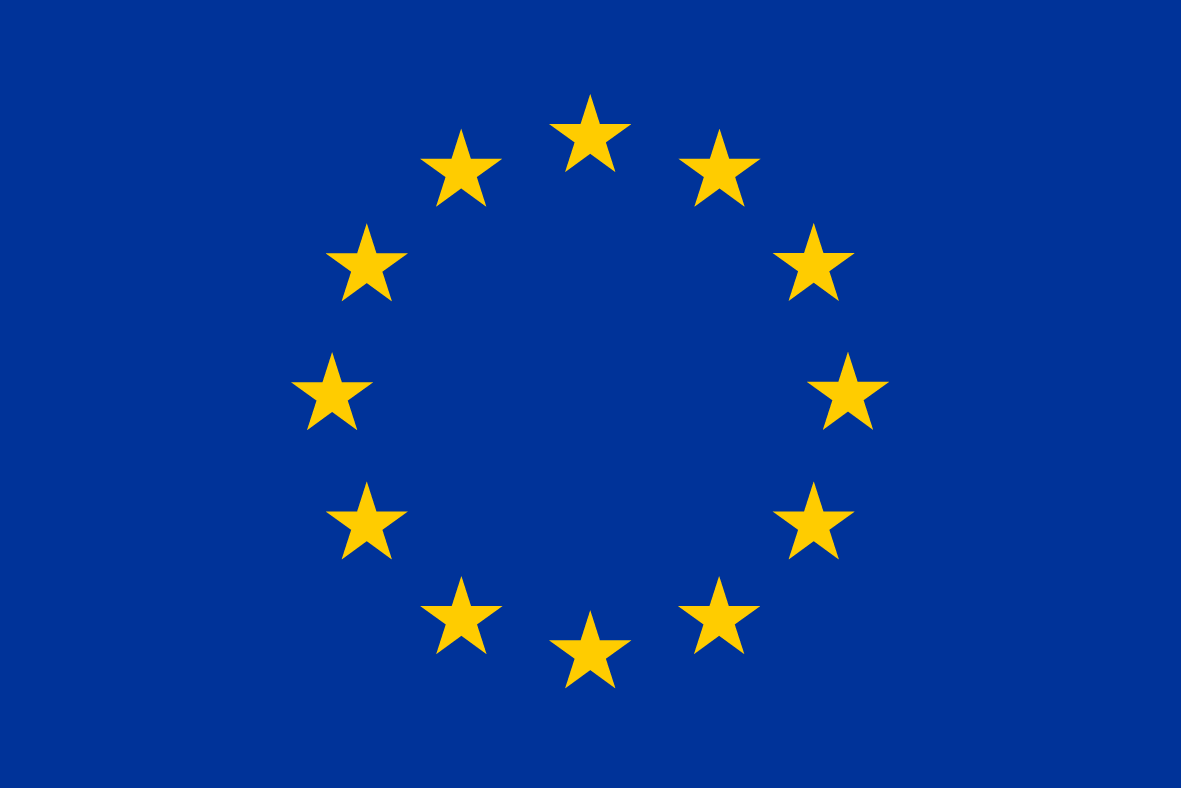}
\end{minipage}
\hfill ~
\end{center}

Funded by the European Union. Views and opinions expressed are however those of the author(s) only and do not necessarily reflect those of the European Union or the European Commission. Neither the European Union nor the European Commission can be held responsible for them. This work was supported by the European Union’s Horizon Europe research and innovation programme under grant agreement No 101214398 (ELLIOT).
\end{ack}

\medskip
{
\small
\bibliographystyle{plainnat}
\bibliography{refs}
}

\newpage
\appendix
\raggedbottom

\crefalias{section}{appendix}
\crefalias{subsection}{subappendix}
\crefalias{subsubsection}{subsubappendix}

\counterwithin{figure}{section}
\counterwithin{table}{section}
\counterwithin{equation}{section}

\section{Benchmark datasets details}
\label{app:datasets}

We provide the full benchmark dataset details in \Cref{tab:benchmarks}.

\begin{table}[H]
  \centering
  \caption{Dataset details on the two considered benchmarks.}
  \vspace{1em}
  \small
  \setlength{\tabcolsep}{5pt}
  \begin{tabular}{@{}llrrl@{}}
    \toprule
    & Campaign & $N$ & $d$ & Objective \\
    \midrule
    \multicolumn{5}{@{}l}{\texttt{liang}: 5 campaigns, 50 seeds, Top-5\% \citep{liang2021benchmarking}} \\
    \addlinespace[2pt]
    & Crossed barrel & 600 & 4 & toughness $\uparrow$ \\
    & P3HT           & 178 & 5 & conductivity ($\frac{S}{cm}$) $\uparrow$ \\
    & AgNP           & 164 & 5 & spectral loss $\downarrow$ \\
    & AutoAM         & 100 & 4 & score $\uparrow$ \\
    & Perovskite     &  94 & 3 & instability index $\downarrow$ \\
    \addlinespace[3pt]
    \midrule
    \multicolumn{5}{@{}l}{\texttt{olympus}: 10 campaigns, 100 seeds, best value at \citeauthor{hase2021olympus}'s checkpoints} \\
    \addlinespace[2pt]
    & hplc           & 1386 & 6 & peak area $\uparrow$ \\
    & photo\_pce10   & 1040 & 4 & degradation $\downarrow$ \\
    & photo\_wf3     & 1040 & 4 & degradation $\downarrow$ \\
    & suzuki         &  247 & 4 & yield $\uparrow$ \\
    & fullerenes     &  246 & 3 & product $\uparrow$ \\
    & colors\_bob    &  241 & 5 & difference to green $\downarrow$ \\
    & alkox          &  208 & 4 & conversion $\uparrow$ \\
    & colors\_n9     &  102 & 3 & difference to green $\downarrow$ \\
    & benzylation    &   73 & 4 & impurity $\downarrow$ \\
    & snar           &   66 & 4 & E-factor $\downarrow$ \\
    \bottomrule
  \end{tabular} 
  \label{tab:benchmarks}
\end{table}

\section{Baselines}
\label{app:baselines}

The full list of surrogates can be found in \Cref{tab:surrogates}, while the five acquisition functions are listed in \Cref{tab:acquisitions}.

\begin{table}[H]
\caption{\textbf{Surrogates.} Every arm we consider is one of these surrogates
under one of the rules in \Cref{tab:acquisitions}, and the set is the full cross
product: eleven surrogates $\times$ five rules gives the 55 models we report. \texttt{Random} is not part of that product, but it is scored like
any other arm and is one of the 56 that set the min-max range. Each arm is run on
all 5 \texttt{liang} and all 10 \texttt{olympus} campaigns, at every seed the
protocol declares (50 and 100 respectively). Kernels are given in
\Cref{tab:kernels}; all are anisotropic (ARD) except \texttt{GP-iso}.}
\vspace{1em}
\centering
\small
\setlength{\tabcolsep}{5pt}

\label{tab:surrogates}
\begin{tabular}{@{}lp{8.6cm}@{}}
\toprule
Surrogate & Family \\
\midrule
\texttt{TabPFN}      & tabular foundation model \citep{grinsztajn2026tabpfn3technicalreport} \\
\texttt{TabICL}      & tabular foundation model \citep{qu2026tabiclv2} \\
\addlinespace[3pt]
\texttt{RF}          & random forest, 100 trees, spread across trees \citep{breiman2001random} \\
\texttt{CatBoost}    & boosting, \texttt{RMSEWithUncertainty} + virtual ensembles \citep{prokhorenkova2018catboost} \\
\addlinespace[3pt]
\texttt{GP-ARD}      & Mat\'ern-5/2, ARD \citep{rasmussen2006gaussian}, the reference kernel \\
\texttt{GP-M32}      & Mat\'ern-3/2, ARD \\
\texttt{GP-M12}      & Mat\'ern-1/2, ARD \\
\texttt{GP-RBF}      & squared exponential, ARD \\
\texttt{GP-iso}      & Mat\'ern-5/2, isotropic: one shared lengthscale \\
\texttt{GP-MLP}      & arccosine (MLP) kernel, ARD, non-stationary \\
\texttt{GP-ARD-Warp} & Mat\'ern-5/2 ARD on warped inputs, BoTorch's \citep{balandat2020botorch} Kumaraswamy parameterization of \citet{snoek2014input} \\
\addlinespace[3pt]
\texttt{Random}      & selects uniformly from the un-queried pool \\
\bottomrule
\end{tabular}
\end{table}

\begin{table}[H]
\caption{\textbf{GP kernels.} The seven kernels of \Cref{tab:surrogates}, written as
$k(\mathbf{x},\mathbf{x}')/\sigma^2$, where $\sigma^2$ is the output scale.
All kernels except \texttt{GP-iso} use one lengthscale per input,
$r^2 = \sum_{i=1}^{d} (x_i - x'_i)^2 / \ell_i^2$; \texttt{GP-iso} uses one shared $\ell$.
Each GP is a BoTorch \texttt{SingleTaskGP} \citep{balandat2020botorch}.
Inputs are scaled to the unit cube, and targets are standardized on the observed rows.
Hyperparameters are fit once by maximizing the exact marginal likelihood, with a lower bound on the noise.
If a fit fails, the GP falls back to its prior; we count these failures per kernel and do not drop them.}
\vspace{1em}
\centering
\small
\setlength{\tabcolsep}{5pt}

\label{tab:kernels}
\begin{tabular}{@{}llc@{}}
\toprule
Arm & Kernel & $k(\mathbf{x},\mathbf{x}')/\sigma^2$ \\
\midrule
\texttt{GP-ARD}      & Mat\'ern-5/2, ARD        & $\left(1 + \sqrt{5}\,r + \tfrac{5}{3} r^2\right) e^{-\sqrt{5}\,r}$ \\[3pt]
\texttt{GP-M32}      & Mat\'ern-3/2, ARD        & $\left(1 + \sqrt{3}\,r\right) e^{-\sqrt{3}\,r}$ \\[3pt]
\texttt{GP-M12}      & Mat\'ern-1/2, ARD        & $e^{-r}$ \\[3pt]
\texttt{GP-RBF}      & squared exponential, ARD & $e^{-r^2/2}$ \\[3pt]
\texttt{GP-iso}      & Mat\'ern-5/2, isotropic  & $\left(1 + \sqrt{5}\,r + \tfrac{5}{3} r^2\right) e^{-\sqrt{5}\,r}$, one $\ell$ \\[3pt]
\texttt{GP-MLP}      & arccosine (MLP)          & $\dfrac{2}{\pi} \arcsin \dfrac{\mathbf{x}^\top W \mathbf{x}' + b}{\sqrt{(\mathbf{x}^\top W \mathbf{x} + b + 1)(\mathbf{x}'^\top W \mathbf{x}' + b + 1)}}$ \\[9pt]
\texttt{GP-ARD-Warp} & Mat\'ern-5/2 ARD, warped & $k_{5/2}\!\left(w(\mathbf{x}), w(\mathbf{x}')\right)$, \; $w_i(x) = 1 - (1 - x_i^{a_i})^{b_i}$ \\
\bottomrule
\end{tabular}
\end{table}

\begin{table}[H]
\caption{\textbf{Acquisition rules.} Each is maximized over the un-queried pool
against the incumbent $f^\star = \max y_\mathrm{obs}$, with ties broken uniformly
on their own random stream rather than by \texttt{argmax}, which on a pool in
file order would score an arm that cannot separate its candidates as a systematic
traversal. Written once here, they apply to every surrogate in
\Cref{tab:surrogates}. For TFMs, the confidence bound is the native predictive
quantile $F^{-1}(\Phi(\beta))$, where $F$ is the predictive cumulative distribution
function and $\Phi$ is the standard normal cumulative distribution function.
For Gaussian predictions this equals $\mu+\beta\sigma$.
\texttt{UCB1} and \texttt{UCB2} differ only in this weight, $\beta=1$ and $\beta=2$; the larger $\beta$ gives more weight to predictive uncertainty and so explores more.
The last column gives the Gaussian forms used by the GP and tree arms;
TabPFN uses its bar distribution and TabICL its predicted quantiles.}
\vspace{1em}
\centering
\small
\setlength{\tabcolsep}{6pt}
\label{tab:acquisitions}
\begin{tabular}{@{}lllp{4.3cm}@{}}
\toprule
Rule & & Definition & Gaussian form, $z = (\mu - f^\star)/\sigma$ \\
\midrule
\texttt{greedy} & \citep{death2021greed}      & $\mu$                                  & $\mu$ \\
\texttt{EI}     & \citep{mockus1978application,jones1998efficient} & $\mathbb{E}\big[(f - f^\star)_+\big]$  & $(\mu - f^\star)\Phi(z) + \sigma\phi(z)$ \\
\texttt{PI}     & \citep{kushner1964new}      & $\Pr\big[f > f^\star\big]$             & $\Phi(z)$ \\
\texttt{UCB1}   & \citep{cox1992statistical,srinivas2010gaussian} & $F^{-1}(\Phi(1))$       & $\mu + \sigma$ \\
\texttt{UCB2}   & \citep{cox1992statistical,srinivas2010gaussian} & $F^{-1}(\Phi(2))$       & $\mu + 2\sigma$ \\
\bottomrule
\end{tabular}
\end{table}

\subsection{Implementation details}
\label{app:impl}

All arms run in \texttt{src/matbo/methods/} of the code repository.

\paragraph{Inputs.} Every surrogate receives the same matrix of the benchmark's numeric input variables. On \texttt{liang}, repeated measurements of one design point become one candidate whose target is the mean of the distinct measured values, following \citet{liang2021benchmarking}; on \texttt{olympus}, every recorded row is a candidate. Minimization targets are negated, so every surrogate maximizes.

\paragraph{TabPFN.} A new TabPFN-3 regressor per round, with 8 estimators, float32 inference and the seed as random state. One fit and one forward pass per round return the bar distribution; the mean, EI, PI and UCB are that distribution's own functions.

\paragraph{TabICL.} TabICLv2 with 8 estimators and the seed as random state. EI is $\int_0^1 \max(Q(u)-f^\star,0)\,du$ over the predicted quantile function $Q$, by the trapezoid rule on levels refined in the upper tail. PI is $1-F(f^\star)$, with $F(f^\star)$ found by interpolating $Q$. UCB is $Q(\Phi(\beta))$. If no feature varies in the context, which can happen when the two initial \texttt{olympus} rows are duplicates, the prediction is flat and the tie-break decides.

\paragraph{GPs.} See \Cref{tab:kernels}. EI is computed as log-EI, which has the same maximizer and does not underflow; PI and UCB use the Gaussian posterior.

\paragraph{Trees.} RF uses 100 trees; its mean and standard deviation are taken across trees. CatBoost uses 200 iterations with the \texttt{RMSEWithUncertainty} loss and 10 virtual ensembles; its standard deviation combines the knowledge and data variance. If the context targets are constant, CatBoost cannot fit, so the prediction is flat and the tie-break decides. Both use the Gaussian forms in \Cref{tab:acquisitions}.

\section{Normalized improvement metric}
\label{app:ni}
The underlying metrics of each benchmark cannot be directly averaged over campaigns as pool sizes range from 66 to 1386.
We therefore normalize \emph{within} a campaign before aggregating. Let $y_a(c,b)$ be
arm $a$'s metric (Top-5\% or best value) on campaign $c$ at budget $b$, averaged over seeds, and $\mathcal{A}$ the set of arms run on every campaign of that benchmark. Then
\begin{equation}
  \mathrm{NI}_a(c,b)=\frac{y_a(c,b)-\min_{a'\in\mathcal{A}}y_{a'}(c,b)}
                          {\max_{a'\in\mathcal{A}}y_{a'}(c,b)-\min_{a'\in\mathcal{A}}y_{a'}(c,b)},
  \qquad
  \overline{\mathrm{NI}}_a(b)=\frac{1}{|\mathcal{C}|}\sum_{c\in\mathcal{C}}\mathrm{NI}_a(c,b),
  \label{eq:ni}
\end{equation}
so the field's worst arm is $0$ and its best is $1$. Both extremes are taken over $\mathcal{A}$ rather than over the arms plotted, so the scale does not move with the choice of what to show. Figure bands $\overline{\mathrm{NI}}_a(b)$ are given by a percentile bootstrap over $\mathcal{C}$ (400 resamples); seeds are averaged inside $y_a(c,b)$ and carry no band.

\section{Additional results}

\subsection{Per-campaign results}
\label{app:raw_results}

\begin{figure}
    \centering
    \includegraphics[width=\linewidth]{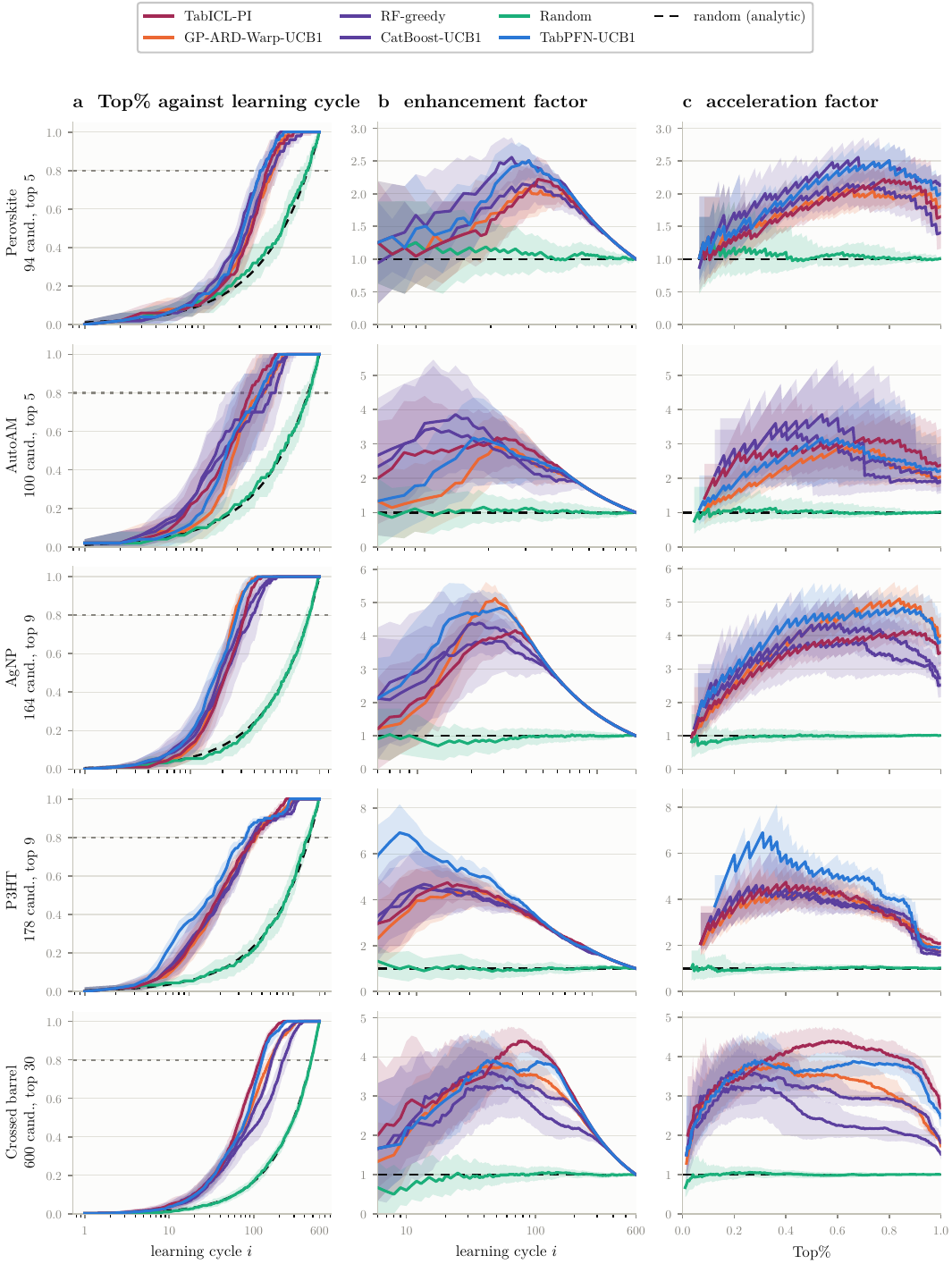}
    \caption{\textbf{Raw per-campaign results on \texttt{liang}}, one row per campaign, the arms of \Cref{fig:ni-by-budget}. \textbf{a}, Top-5\% found against learning cycle. \textbf{b}, enhancement factor and \textbf{c}, acceleration factor, both as defined by \citet{liang2021benchmarking}, against a random baseline of $1$ (dashed). Bands are the $5$--$95$ percentile over fold means (cf. \citet{liang2021benchmarking} for full details).}
    \label{fig:liang_raw}
\end{figure}

\begin{figure}
    \centering
    \includegraphics[width=\linewidth]{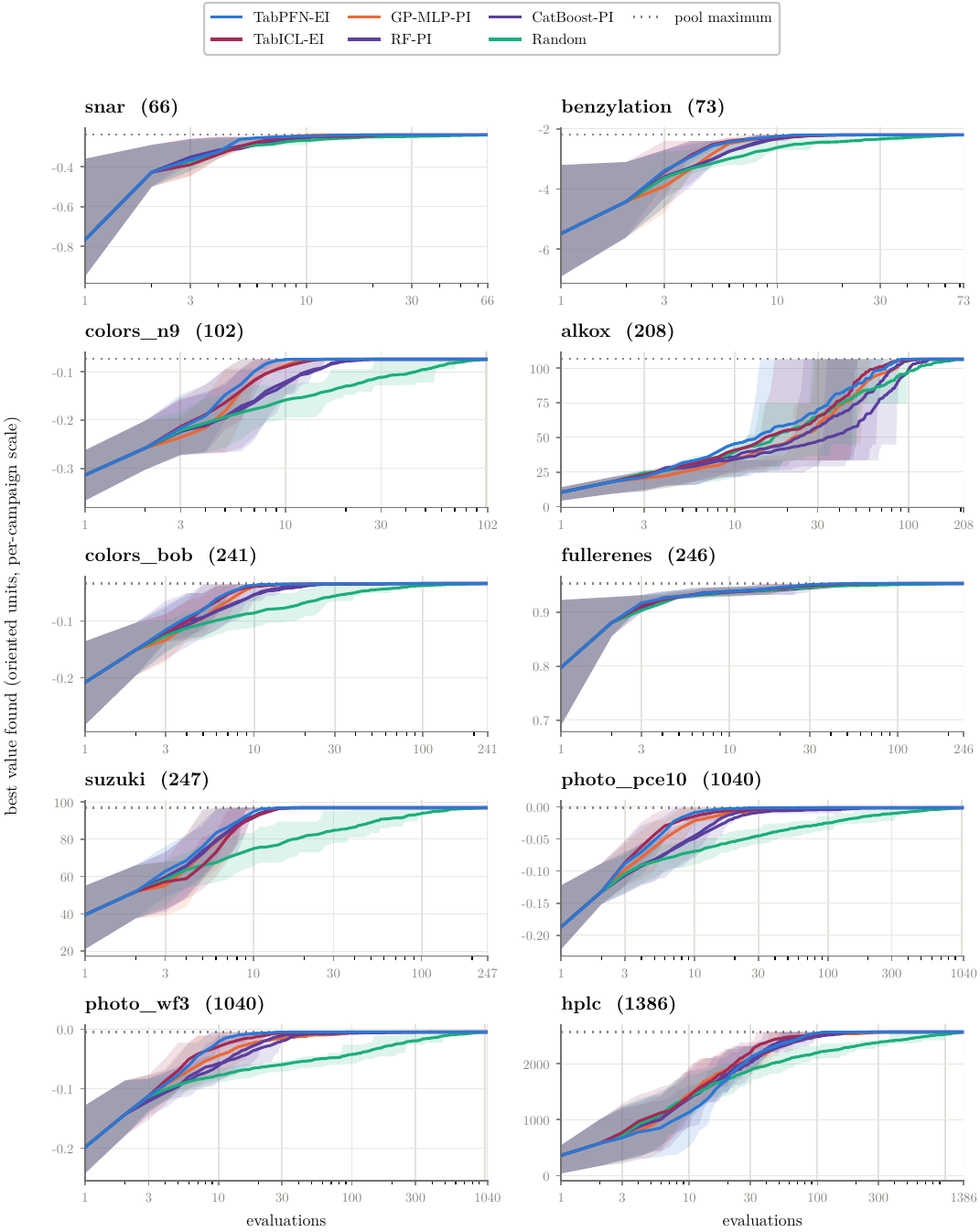}
    \caption{\textbf{Raw per-campaign results on \texttt{olympus}}, one panel per campaign with pool size in brackets, the arms of \Cref{fig:ni-by-budget}. Best value found against number of evaluations, on each campaign's own oriented scale; the dotted line is the pool maximum. Bands are the $25$--$75$ percentile over 100 runs, reported following the protocol of \citet{hase2021olympus}.}
    \label{fig:olympus_raw}
\end{figure}

\begin{figure}
    \centering
    \includegraphics[width=0.97\linewidth]{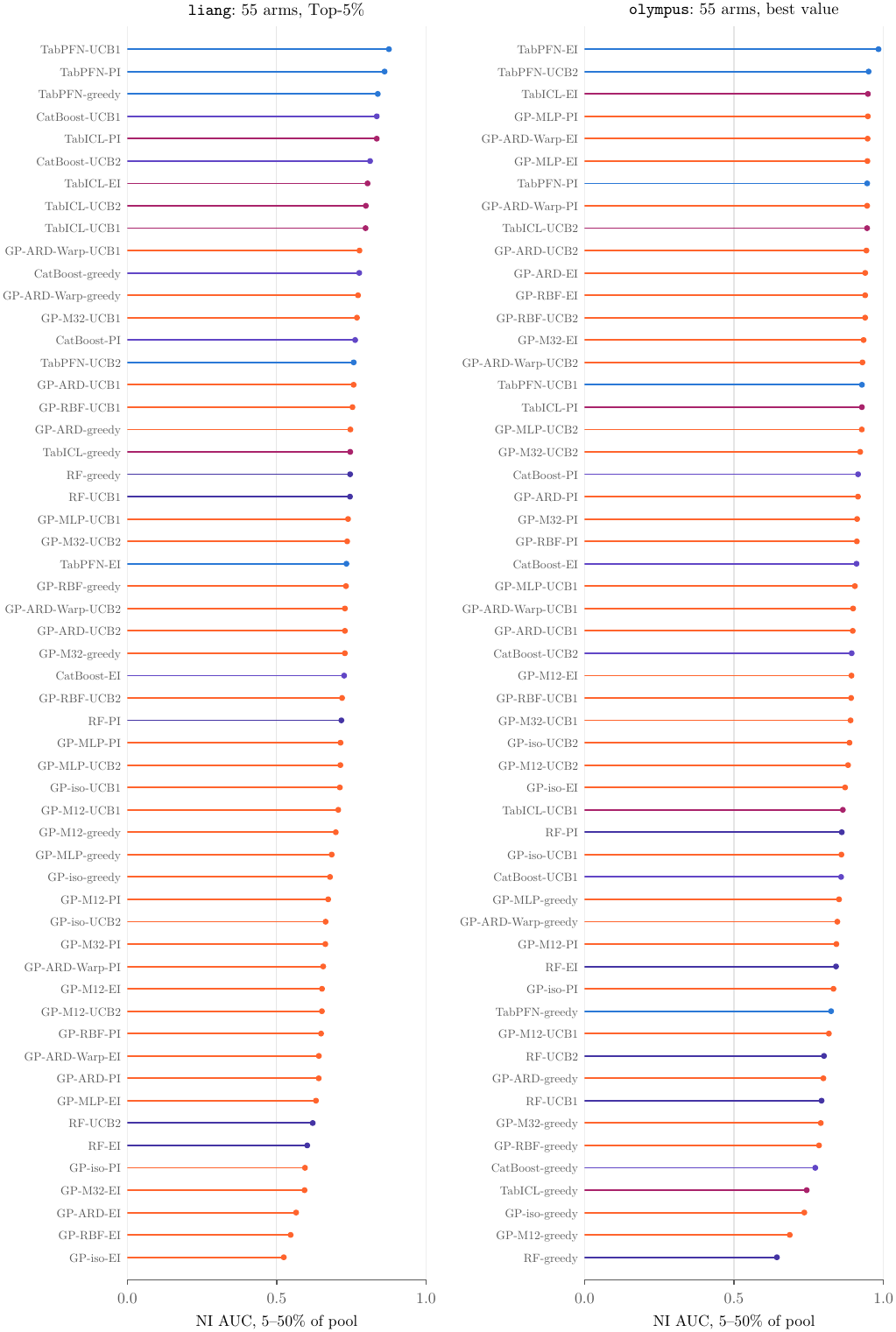}
     \caption{%
      \textbf{Every arm on both benchmarks, ranked by normalized improvement.} Each arm's NI-versus-budget curve (cf. \Cref{fig:all_arms_liang_bestvalue,fig:all_arms_olympus_top5}) is collapsed to one number: the integral of NI over budget from 5\% to 50\% of the pool, divided by the length of that axis so it reads on the NI scale. Color represents the surrogate family. \texttt{Random} enters the normalization as an ordinary arm but is omitted from the ranking. Each benchmark is scored on the metric its own paper reports: (a)~\texttt{liang}  on Top-5\% found; (b)~\texttt{olympus} on best achieved value.%
    }
    \label{fig:ni_auc_bars_main}
\end{figure}

\begin{figure}[h]
    \centering
    \includegraphics[width=\linewidth]{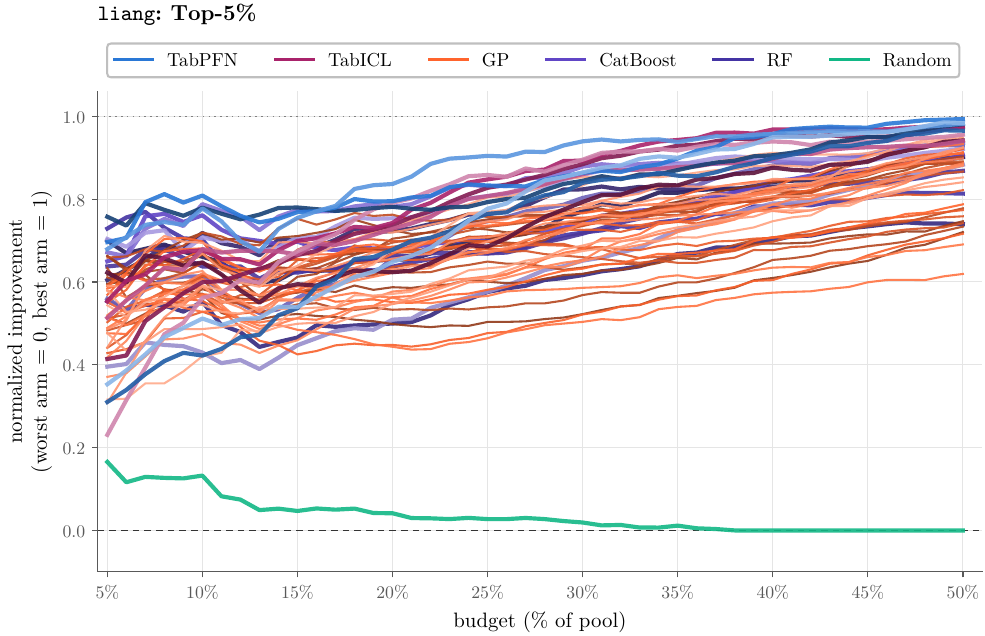}
     \caption{\textbf{Full set of all 56 methods on \texttt{liang}, scored on Top-5\%.} Color marks the surrogate family.%
    }
    \label{fig:all_arms_liang_bestvalue}
\end{figure}

\begin{figure}[h]
    \centering
    \includegraphics[width=\linewidth]{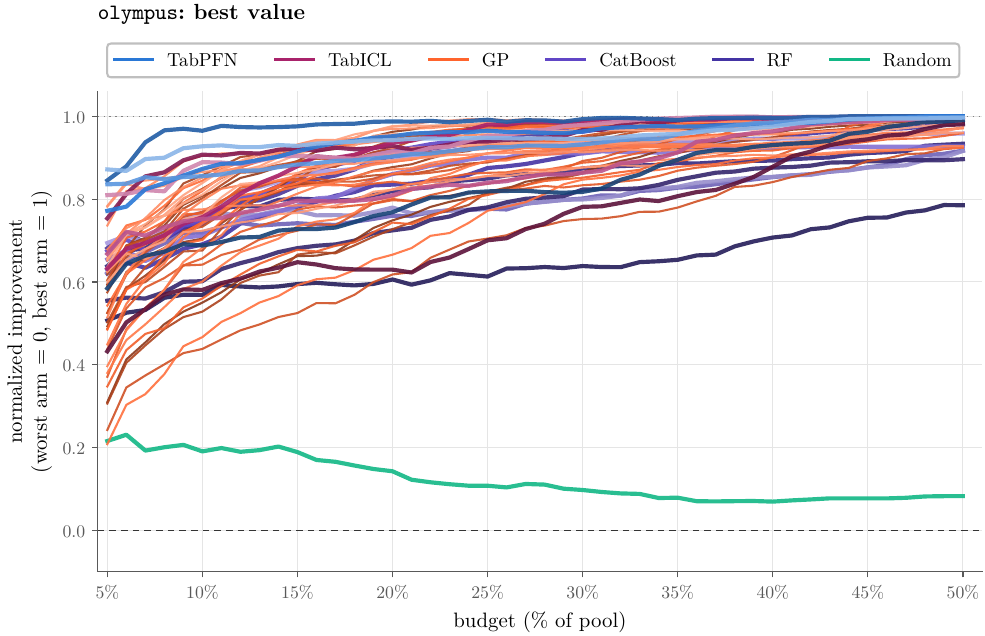}
     \caption{\textbf{Full set of all 56 methods on \texttt{olympus}, scored on best achieved value.} Color marks the surrogate family.}
    \label{fig:all_arms_olympus_top5}
\end{figure}

Here we show the per-campaign results behind the aggregate figures in the main text (\Cref{fig:ni-by-budget}), following the visualization scheme of each benchmark's original paper \citep{liang2021benchmarking,hase2021olympus}. \Cref{fig:liang_raw} shows the five \texttt{liang} campaigns as Top-5\% against learning cycle together with the enhancement and acceleration factors of \citet{liang2021benchmarking}. \Cref{fig:olympus_raw} shows the best value found against evaluations on each of the ten \texttt{olympus} campaigns.
\Cref{fig:ni_auc_bars_main} shows each benchmark's ranking of every method, according to the area under the normalized improvement curve. This ranking is the basis upon which the selection of each surrogate model family's representative for plotting is performed. The underlying NI curves for all methods are presented in \Cref{fig:all_arms_liang_bestvalue,fig:all_arms_olympus_top5}.

\clearpage

\subsection{Metric dependence}
\label{app:metric_flip}

\begin{figure}
    \centering
    \includegraphics[width=\linewidth]{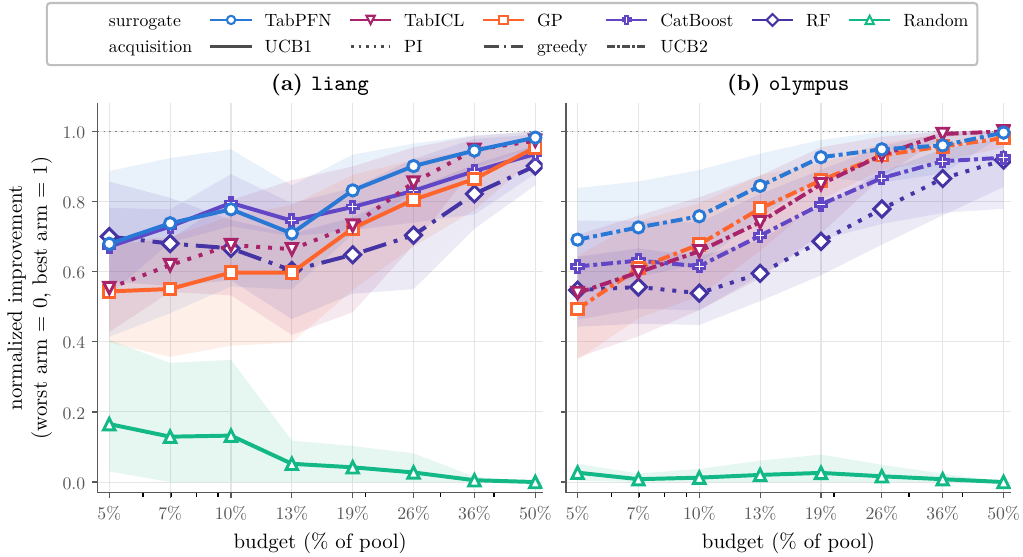}
    \caption{\textbf{Both benchmarks scored on Top-5\% found.} Normalized improvement against budget, best arm per surrogate family. Panel~(a) repeats the main-text scoring, panel~(b) is the flipped one.}
    \label{fig:ni_topcount_both}
\end{figure}
\begin{figure}
    \centering
    \includegraphics[width=\linewidth]{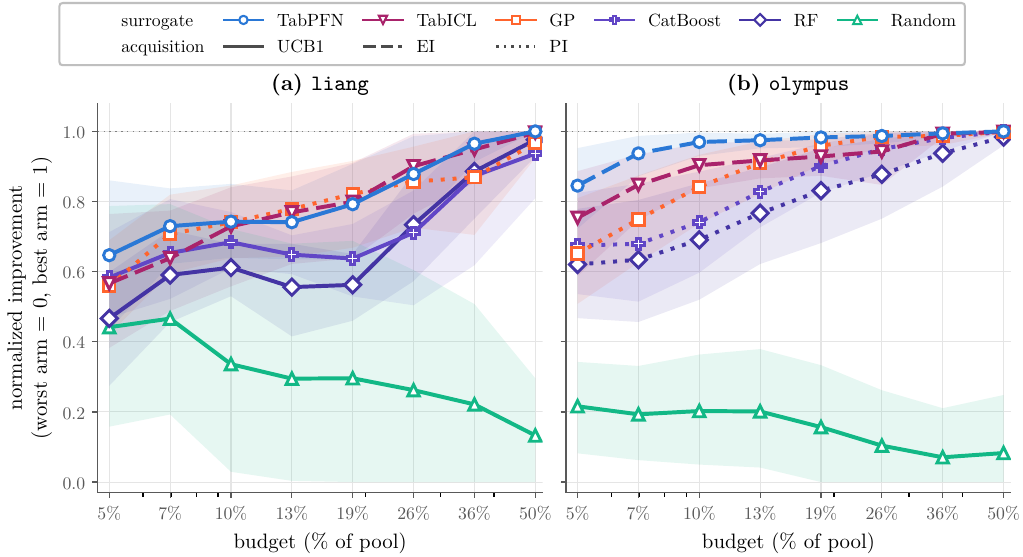}
    \caption{\textbf{Both benchmarks scored on best achieved value.} Normalized improvement against budget, best arm per surrogate family. Panel~(b) repeats the main-text scoring, panel~(a) is the flipped one.}
    \label{fig:ni_bestvalue_both}
\end{figure}

\Cref{fig:ni-by-budget,fig:ni_auc_bars_main} score each benchmark on the metric its own paper reports: Top-5\% found on \texttt{liang}, best achieved value on \texttt{olympus}. Here we score the same runs on the other benchmark's metric. \Cref{fig:ni_topcount_both} puts both benchmarks on Top-5\% found and \Cref{fig:ni_bestvalue_both} puts both on best achieved value, so each metric can be read across benchmarks.

\clearpage
\subsection{GP kernel variation on \texttt{olympus}}
\label{app:kernel_olymp}

\begin{figure}
  \centering
  \includegraphics[width=\linewidth]{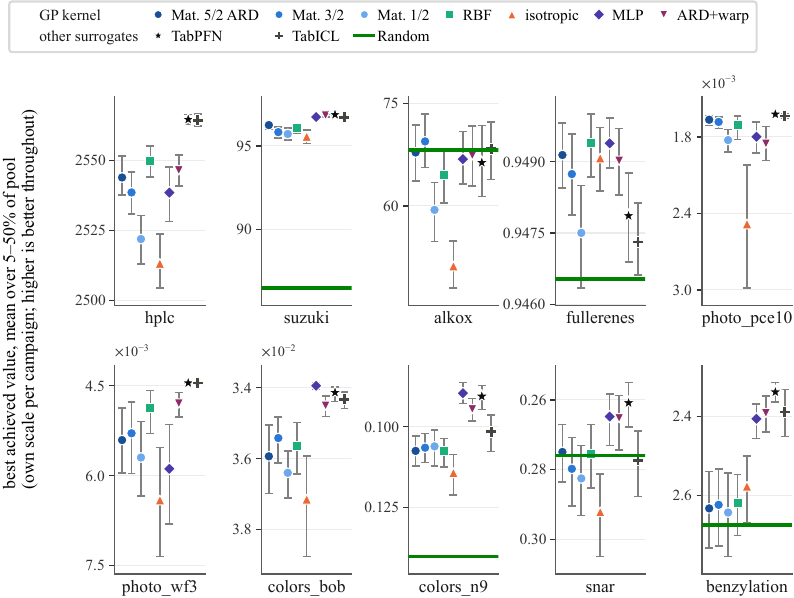}
  \caption{\textbf{GP kernel choice per dataset on \texttt{olympus}.} All arms under
    \textsc{pi},  the acquisition rule of the best-ranked GP arm on that benchmark, in the benchmark's declared metric. Each mark is one arm's best-value curve averaged over eight log-spaced budgets from 5-50\% of the candidate pool; bars are a $2.5$-$97.5$ percentile bootstrap over the protocol's 100 seeds, taken on each seed's own average. Every campaign has its own axis, since best value has no common scale here, and is shown in the source's own units.}
  \label{fig:kernels_olympus_pi}
\end{figure}

\Cref{fig:kernels_olympus_pi} shows the per-dataset kernel comparison on
\texttt{olympus} under \textsc{pi}, the acquisition rule of the best-ranked GP arm on that benchmark, analogous to \Cref{fig:kernels_ucb2} for \texttt{liang}. Six of the ten campaigns are minimizations: \texttt{benzylation} (impurity), \texttt{snar} (E-factor), \texttt{colors\_bob} and \texttt{colors\_n9} (difference to green), and \texttt{photo\_pce10} and \texttt{photo\_wf3} (degradation). Their axes are inverted, so better is upward in every panel. Star: TabPFN; cross: TabICL; green line: random search, omitted from \texttt{hplc}, \texttt{colors\_bob}, \texttt{photo\_pce10} and \texttt{photo\_wf3}, where it lies far enough below the field to compress every arm into a band. Datasets are ordered by the median kernel.

\clearpage
\subsection{Seed-level tests and campaign-specific acquisition choice}
\label{app:posthoc}

We analyze the saved trajectories over the same linear 5--50\% budget grid as \Cref{fig:ni_auc_bars_main}. For each seed, performance AUC integrates the raw campaign metric and divides by the window width, giving Top-5\% points on \texttt{liang} and objective units on \texttt{olympus}.

\Cref{tab:seed_tests} tests the arms of \Cref{fig:ni-by-budget} within each campaign: the TabPFN arm against each other family's arm, which may use a different acquisition rule. As in \citet{shields2021bayesian}, a Welch t-test compares the mean performance AUC over seeds. Both arms are selected on the evaluated results, so the counts do not correct for that selection.

Across campaigns, we rank the six families within each campaign by mean performance AUC and apply a Friedman test with the Nemenyi post-hoc test \citep{demsar2006statistical}. TabPFN has the best mean rank. At $\alpha=0.05$, it is separated from CatBoost, RF and Random, but not from TabICL or GP. The GP arm is the best of 35 configurations, chosen post hoc. Ranked by median performance AUC instead, TabPFN is separated only from Random: on four \texttt{olympus} campaigns, most families reach the best candidate in more than half of their runs, so their medians tie.

\begin{table}[H]
  \centering
  \caption{\textbf{Seed-level tests of \Cref{fig:ni-by-budget}.} Number of campaigns on which a Welch t-test ($\alpha=0.05$) on performance AUC over seeds finds the TabPFN arm better or worse than the comparator, or does not separate them. Arms are those of \Cref{fig:ni-by-budget}: TabPFN-UCB1 on \texttt{liang} and TabPFN-EI on \texttt{olympus}.}
  \vspace{0.5em}
  \small
  \begin{tabular}{@{}llccc@{}}
    \toprule
    Benchmark & Comparator & better & worse & not separated \\
    \midrule
    \texttt{liang} (5)    & TabICL-PI        & 3 & 1 & 1 \\
                          & GP-ARD-Warp-UCB1 & 3 & 0 & 2 \\
                          & CatBoost-UCB1    & 3 & 0 & 2 \\
                          & RF-greedy        & 4 & 0 & 1 \\
    \addlinespace[2pt]
    \texttt{olympus} (10) & TabICL-EI        & 6 & 0 & 4 \\
                          & GP-MLP-PI        & 7 & 1 & 2 \\
                          & CatBoost-PI      & 7 & 0 & 3 \\
                          & RF-PI            & 9 & 0 & 1 \\
    \bottomrule
  \end{tabular}
  \label{tab:seed_tests}
\end{table}

\clearpage
\paragraph{Aggregate ranking versus campaign-specific acquisition choice.}
\Cref{fig:ni_auc_bars_main} ranks surrogate--acquisition configurations by NI-AUC, which normalizes performance within each campaign before averaging across campaigns and integrating over budget. \Cref{fig:ni-by-budget} uses this ranking to select one configuration per surrogate family for each benchmark; it does not select a different configuration for each campaign.

\begin{figure}[H]
  \centering
  \includegraphics[width=0.74\linewidth]{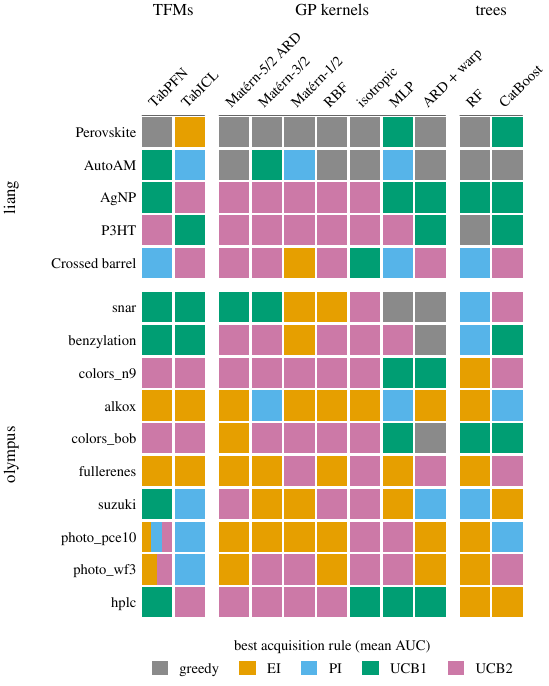}
  \caption{\textbf{Best observed acquisition rule by surrogate and campaign.} Each cell identifies the rule with the highest mean raw-metric AUC over the 5--50\% budget window. Stripes indicate exact ties. Unlike \Cref{fig:ni_auc_bars_main}, this figure makes no comparison aggregated across campaigns. Winners are selected on the evaluated runs; a winning rule need not be statistically distinguishable from the alternatives. Colors identify rules, not the size of their advantage.}
  \label{fig:posthoc_best_rule}
\end{figure}

The best-rule map in \Cref{fig:posthoc_best_rule} answers a different question: which acquisition rule performs best for each surrogate on each individual campaign? It selects the rule with the highest mean raw-metric AUC, without cross-campaign normalization or averaging. Consequently, the best aggregate configuration need not win on every campaign. For example, TabPFN-UCB1 leads the \texttt{liang} NI-AUC ranking but is the best TabPFN acquisition on only two of its five campaigns.

\clearpage
\section{Differences with \texorpdfstring{\citet{hu2026foundation}}{Hu et al.}}
\label{sec:hu}

\citet{hu2026foundation} also study the performance of TFMs as surrogates for BO in materials science. Here we provide more details on our differences with this paper.

\paragraph{Different evaluation datasets.} \citet{hu2026foundation} assemble their own collection of ten datasets (copper alloys, bulk metallic glasses, lattice thermal conductivity; 495--3,148 candidates) and score methods on a custom metric, the number of extra evaluations to reach the global optimum, starting from an initial set chosen as the bottom-$K\%$ of samples by property value. We instead evaluate on two previously published benchmarks \citep{liang2021benchmarking, hase2021olympus} (15 campaigns, 66--1386 candidates), following each benchmark's own protocol, seeds, and metric, with a random two-point initial design.

\paragraph{More baselines.} \citet{hu2026foundation} compare TabPFN against a single GP and a random forest. We report the full cross product of eleven surrogates and five acquisition rules (55 arms, plus random search), including seven GP kernels, CatBoost, and a second tabular foundation model (TabICL), which lets us separate the effect of the surrogate family from that of the kernel or acquisition choice. In addition, we always vary the random seed for all models compared, including TabPFN, whereas \citet{hu2026foundation} do not vary it for TabPFN.

\paragraph{Taking advantage of TabPFN's native uncertainty quantification.} \citet{hu2026foundation} use a Gaussian approximation for the uncertainty of TabPFN, and introduce normalized and hybrid acquisition functions, in particular pairing TabPFN's mean with a GP's variance, to compensate for perceived weaknesses of TabPFN's uncertainty estimates. We use TabPFN and TabICL strictly out-of-the-box, integrating standard acquisition rules directly against their native predictive distributions.

\paragraph{Differences in conclusions.} While we share the overall conclusion that TFMs are strong surrogates, \citet{hu2026foundation} report a minimum-data threshold of roughly 10--20\% of the pool before TabPFN overtakes a GP. We do not observe this: in our setting the best TabPFN arm is already ahead of the best GP arm at the smallest budget we report (5\% of the pool), on both benchmarks (\Cref{fig:ni-by-budget}).

\end{document}